\documentclass[conference]{ieeeconf}  

\IEEEoverridecommandlockouts                              

\usepackage{cite}
\usepackage{xcolor,soul}
\usepackage{tcolorbox} 
\soulregister\cite7
\soulregister\citep7
\soulregister\citet7
\soulregister\ref7
\soulregister\pageref7

\usepackage{amssymb}
\usepackage{bm}
\usepackage{flushend}

\usepackage{multirow}
\usepackage{soul}
\usepackage{svg}
\usepackage{float}

\usepackage{hyperref}
\usepackage{booktabs}

\hypersetup{
    colorlinks=true,
    linkcolor=black,
}

\usepackage{cuted}
\usepackage{capt-of}
\usepackage{caption}
\usepackage{amsmath}

\usepackage{blindtext}
\usepackage{tabularx}
\usepackage{cuted}
\usepackage{amsmath}

\title{\LARGE \bf
Adaptive Motorized LiDAR Scanning Control for Robust Localization with OpenStreetMap}

\author{Jianping Li*,~\IEEEmembership{Member,~IEEE}, Kaisong Zhu*, Zhongyuan Liu, \\ Rui Jin, Xinhang Xu, Pengfei Wan and Lihua Xie,~\IEEEmembership{Fellow,~IEEE}}

\begin{document}

\twocolumn[{%
\renewcommand\twocolumn[1][]{#1}%
\begin{center}
    \centering
    \captionsetup{type=figure}
    \maketitle
    \includegraphics[width=\textwidth]{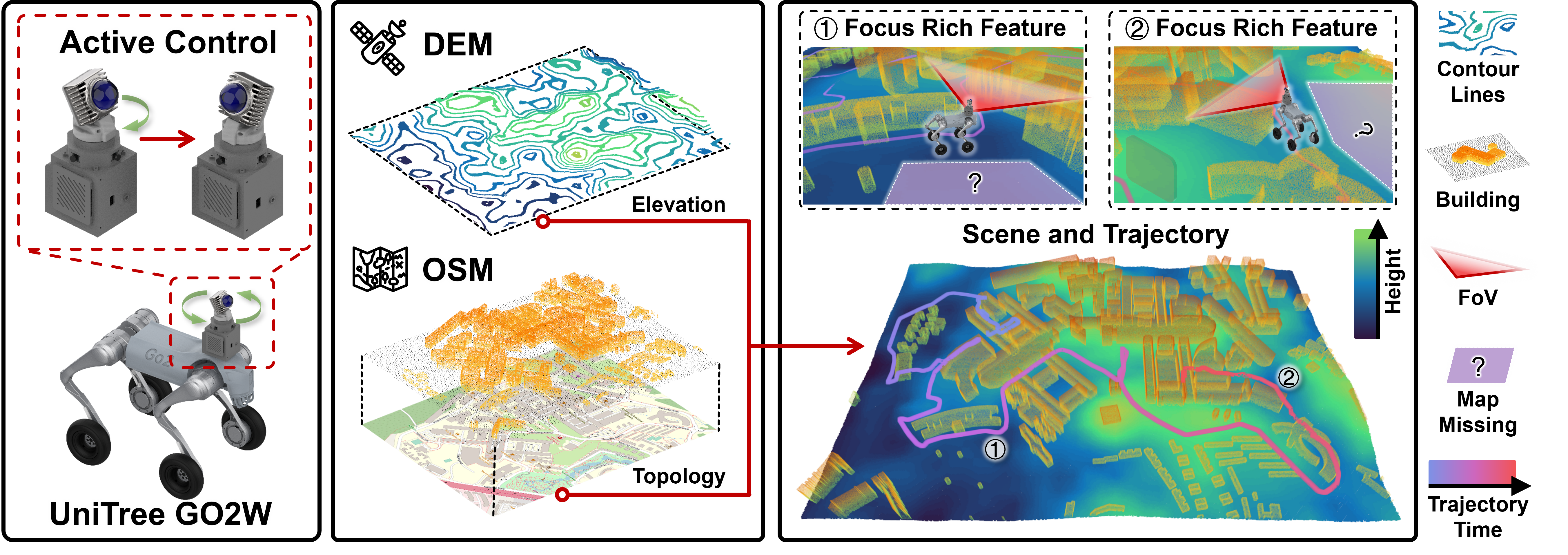}
    \captionof{figure}{Overview of the proposed OSM-guided adaptive LiDAR scanning framework. 
Our method couples local observability prediction with lightweight global priors from OSM and Shuttle Radar Topography Mission (SRTM), enabling the LiDAR to dynamically allocate scanning effort toward informative directions. 
This joint design addresses the limitations of constant-speed rotation and incomplete map data, 
providing robust and efficient localization in complex environments.}
    \label{fig:abstract}
\end{center}%
}]

\renewcommand{\thefootnote}{}
\footnotetext{This work was supported by NTUitive Gap Fund (NGF-2025-17006) and the National Research Foundation, Singapore, under its Medium-Sized Center for Advanced Robotics Technology Innovation. J. Li, K. Zhu, Z. Liu, R. Jin, X. Xu, P. Wan and L. Xie are with School of Electrical and Electronic Engineering, Nanyang Technological University, 50 Nanyang Avenue, Singapore.  (E-mail:jianping.li@ntu.edu.sg,kzhu005@e.ntu.edu.sg, zliu051@e.ntu.edu.sg,bbbbigrui@gmail.com,XU0021NG@e.ntu.edu.sg,
pengfei008@e.ntu.edu.sg,elhxie@ntu.edu.sg) (Jianping Li and Kaisong Zhu contribute equally to this work.)}


\begin{abstract}

LiDAR-to-OpenStreetMap (OSM) localization has gained increasing attention, as OSM provides lightweight global priors such as building footprints. These priors enhance global consistency for robot navigation, but OSM is often incomplete or outdated, limiting its reliability in real-world deployment. Meanwhile, LiDAR itself suffers from a limited field of view (FoV), where motorized rotation is commonly used to achieve panoramic coverage. Existing motorized LiDAR systems, however, typically employ constant-speed scanning that disregards both scene structure and map priors, leading to wasted effort in feature-sparse regions and degraded localization accuracy. To address these challenges, we propose Adaptive LiDAR Scanning with OSM guidance, a framework that integrates global priors with local observability prediction to improve localization robustness. Specifically, we augment uncertainty-aware model predictive control with an OSM-aware term that adaptively allocates scanning effort according to both scene-dependent observability and the spatial distribution of OSM features. The method is implemented in ROS with a motorized LiDAR odometry backend and evaluated in both simulation and real-world experiments. Results on campus roads, indoor corridors, and urban environments demonstrate significant reductions in trajectory error compared to constant-speed baselines, while maintaining scan completeness. These findings highlight the potential of coupling open-source maps with adaptive LiDAR scanning to achieve robust and efficient localization in complex environments.

\end{abstract}

\section{Introduction}


Accurate and robust localization is fundamental for autonomous robots operating in large-scale environments \cite{zou2025reliable,li2025aeos,li2025saliencyi2ploc,yang2022hierarchical}. Beyond relying solely on onboard sensing, recent studies show that leveraging open-source geographic data such as OpenStreetMap (OSM) can provide valuable priors for localization \cite{cho2022openstreetmap}. By aligning LiDAR observations with road centerlines and building footprints extracted from OSM, robots can enforce global consistency in pose estimation, reduce long-term drift, and lower the dependency on dense pre-built maps \cite{li2025graph}. In practice, however, OSM is often incomplete or missing fine-grained structures (e.g., narrow service roads, small courtyards, façade details)\cite{liao2024osmloc}. These issues introduce spatially varying prior quality and can cause pose ambiguity where the OSM geometry is sparse or outdated. As a result, OSM alone is insufficient as a stand-alone reference and must be complemented with sensing strategies that explicitly account for the availability and reliability of map priors.

Meanwhile, LiDAR has established itself as a cornerstone modality for robotic localization due to its high-fidelity 3D measurements and robustness to illumination and texture variations \cite{SU2021103759}. Yet, a single LiDAR inherently suffers from a limited field of view (FoV) \cite{li2025limo} \cite{Kim_2024}: at any instant, only a narrow angular sector is observed, which exacerbates degenerate geometries such as long corridors, parallel façades, and open roads with few salient structures. To mitigate this, motorized LiDAR systems rotate the sensor to achieve panoramic coverage without adding multiple scanners. However, most existing systems adopt constant-speed rotation, which ignores both environmental structure and the spatial distribution of usable map priors. Consequently, the scan budget is spent uniformly, leading to over-sampling of uninformative directions and under-sampling of informative ones; localization accuracy then deteriorates in feature-sparse sectors and in regions where OSM constraints are weak or absent \cite{li2025ua}. These observations suggest that rotation should be treated as a controllable resource to be allocated adaptively rather than a fixed pattern to be executed blindly.

These challenges motivate a new paradigm: \emph{adaptive LiDAR scanning guided by OSM priors}. The key idea is to couple local, sensor-driven observability with global, map-driven guidance \cite{10.3389/frobt.2023.1064934} so that the motor allocates more dwell time and angular resolution toward directions that (i) increase LiDAR odometry observability and (ii) intersect OSM structures likely to provide discriminative constraints \cite{cho2022openstreetmapbasedlidargloballocalization} when present, while still ensuring sufficient coverage where OSM is missing. By dynamically steering the scanning effort according to both the scene and the spatial distribution of OSM features, the system can maintain odometry accuracy in feature-sparse views \cite{isprs-archives-XLVIII-2-W7-2024-73-2024}, enhance alignment with available map geometry, and gracefully degrade in map-incomplete areas. This joint consideration addresses the limitations of purely sensor-based control (which neglects the global structure) and purely map-based localization (which is brittle to prior incompleteness) \cite{aerospace11020120}, paving the way for robust and efficient localization in complex, real-world environments.

In this work, we present \textit{Adaptive LiDAR Scanning with OSM Guidance}, an OSM-aware motor control framework for robust localization. The main contributions are as follows: 

\begin{itemize}
    \item \textbf{OSM-guided adaptive scanning framework:} We integrate OSM features into the scanning control process, converting road and building layouts into regions of interest (RoIs) that guide scan allocation.  
    \item \textbf{OSM-aware optimization objective:} Building on uncertainty-aware model predictive control, we introduce a prior term that biases scanning toward OSM-relevant regions while preserving efficiency.  
    \item \textbf{Integrated system with real-world validation:} We implement the method in ROS with a motorized LiDAR odometry backend and an OSM parser, and validate it extensively in both simulation and field experiments.  
\end{itemize}  

To the best of our knowledge, this is the first work to couple open-source maps with adaptive LiDAR scanning, explicitly integrating global priors and local observability to achieve robust and efficient localization in complex environments.

\section{Related Works}

\subsection{LiDAR-to-Map Localization with OSM}
OpenStreetMap (OSM) provides lightweight and openly accessible geographic data such as road centerlines and building footprints. These priors have been increasingly exploited to support LiDAR-based localization \cite{1730e2efb376477bb66bcbf54c499818}, reducing reliance on dense pre-built maps and improving global consistency over long trajectories \cite{ninan2022lidardatabasedsegmentation}. For instance, Cho et al.~\cite{cho2022openstreetmap} align LiDAR and OSM by generating angle-based distance descriptors and achieve accurate global localization without prior LiDAR maps; likewise, Elhousni et al.~\cite{elhousni2022lidar} employed a constrained particle filter using OSM geometry to maintain sub-3 m tracking accuracy in GPS-denied environments.

Typical approaches match LiDAR measurements to OSM geometry using geometric or semantic constraints (e.g., point-to-line/plane distances, probabilistic map matching), demonstrating that sparse open-source maps can provide meaningful global priors. Recent advances such as OPAL~\cite{kang2025opal} introduce visibility-aware cross-modal alignment via adaptive radial fusion, bridging LiDAR–OSM modality gaps and improving place recognition speed and recall.

In practice, however, OSM is often incomplete or outdated, missing fine-grained structures (e.g., narrow service roads, courtyards, temporary constructions). Additional issues include heterogeneous accuracy/resolution, coordinate mismatches \cite{Ma_2015}, and semantic noise. These limitations result in spatially varying prior quality and can introduce pose ambiguities if OSM is treated as the sole reference \cite{KangLiaoXia2025_1000183839}. Consequently, OSM priors are most effective when combined with complementary sensing strategies that can compensate for map deficiencies \cite{article} and leverage the spatial distribution of OSM features to guide sensor behavior—our proposed method leverages exactly this principle to inform adaptive LiDAR scanning.

\subsection{Motorized LiDAR and Adaptive Scanning}
A single LiDAR inherently suffers from a limited field of view (FoV), restricting its usability in large-scale localization and mapping. To overcome this limitation, motorized LiDAR systems rotate the sensor to provide panoramic coverage without requiring multiple scanners. Early systems such as Zebedee demonstrated that mechanically actuated LiDAR can generate dense 3D maps with a compact setup \cite{bosse2012zebedee}. Subsequent work investigated the calibration of rotating LiDARs, including full-DOF extrinsic estimation using plane measurements \cite{kang2016fulldof} and automatic targetless approaches for spinning LiDARs \cite{alismail2015automatic}, as well as improved plane-based calibration methods \cite{zeng2018improved}.  

While these systems guarantee uniform coverage, constant-speed rotation ignores the spatial distribution of environmental features. As a result, scanning resources are often wasted on uninformative directions, and localization performance deteriorates in degenerate scenarios such as corridors, tunnels, or open roads. To address these limitations, Zhen et al.\ proposed to evaluate localization and observability using a rotating laser scanner \cite{zhen2017robust}, and more recent efforts such as LoLa-SLAM introduced low-latency scan slicing for Lissajous-like trajectories, achieving faster updates with rotating LiDARs \cite{karimi2021lola}. These works highlight that the design of rotation patterns and adaptive scanning strategies can significantly affect odometry accuracy and efficiency.  

Nevertheless, existing methods remain purely sensor-centric. They rely on instantaneous LiDAR geometry or designed motion patterns, but do not exploit external priors that indicate where informative structures are likely to exist. In particular, none of these studies incorporate map knowledge such as OSM into the scanning control loop \cite{s22145206}. This motivates our work: by coupling local observability prediction with OSM-derived spatial cues, motorized LiDAR can allocate scanning effort not only based on immediate sensor data but also according to global map priors, achieving both efficiency and robustness in complex environments.  

\section{Method}

\subsection{Problem Statement and Notation}

\begin{figure}[]
\includegraphics[width=\linewidth]{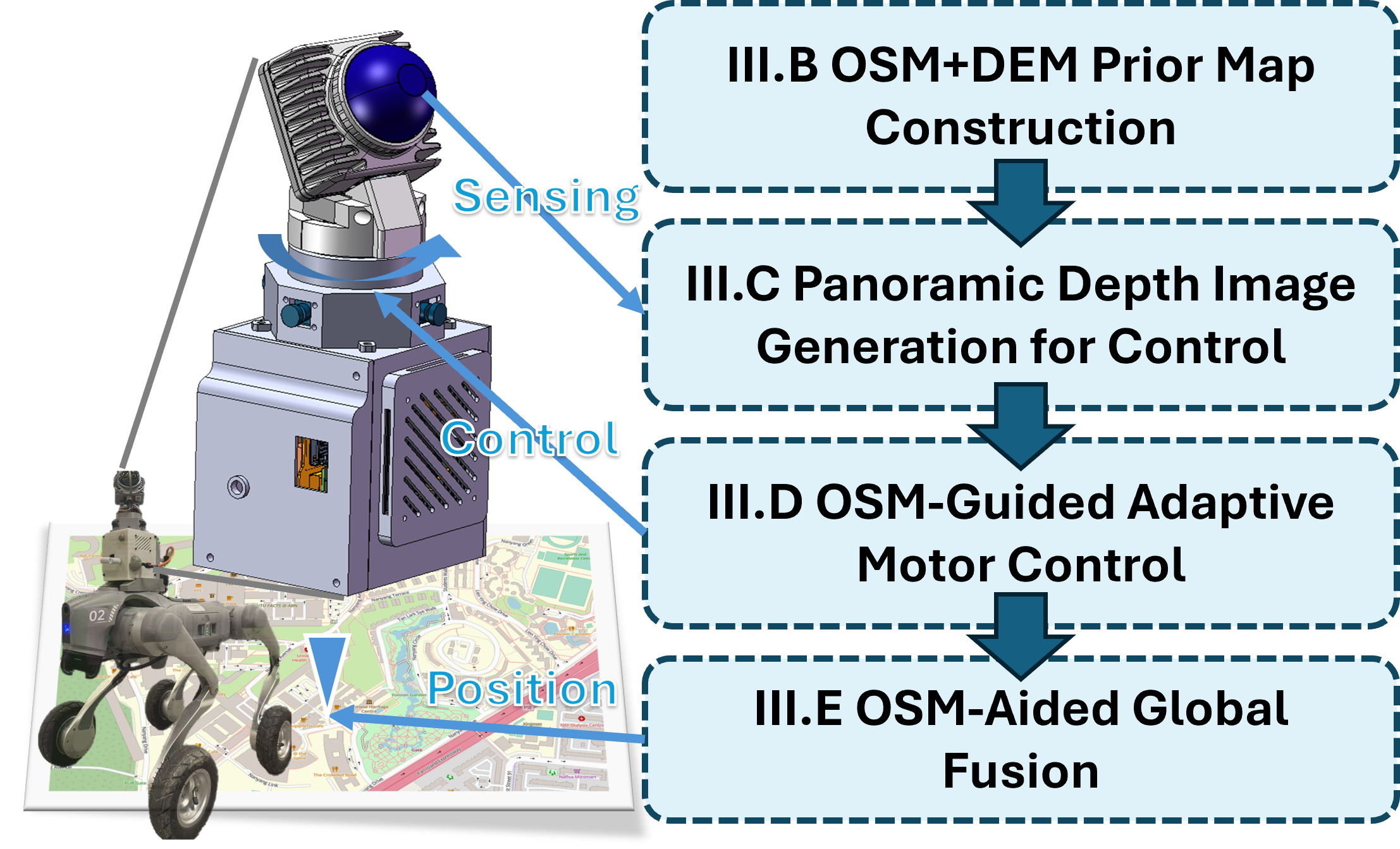}
    \captionof{figure}{Framework for the proposed OSM guided scanning control.}
    \label{fig:framework}
\end{figure}

We summarize the main notation in Table~\ref{tab:notation}. 
Frames are denoted by curly brackets, scalars by italic letters, vectors by bold lowercase, and matrices by bold uppercase. 
For readability, only the most relevant variables are listed here; detailed definitions are given inline in each subsection.

\begin{table}[h]
\centering
\caption{Summary of notation}
\label{tab:notation}
\begin{tabular}{ll}
\hline
Symbol & Definition \\
\hline
$\{L\},\{M_t\},\{B_t\},\{W\}$ & LiDAR, motor, base, world frames \\
$\mathbf{R}^W_B,\mathbf{r}^W_B$ & Base pose in world frame \\
$\mathbf{R}^B_M,\mathbf{R}^M_L$ & Motor and LiDAR extrinsics \\
$\theta,\omega$ & Motor rotation angle and speed \\
$\mathcal{M}_{\text{OSM}}$ & 3D prior map from OSM and DEM \\
$\mathcal{H}_k,\mathcal{V}_k$ & History scans and FoV voxel set \\
$U(\theta)$ & LO uncertainty at orientation $\theta$ \\
$P(\theta)$ & OSM prior utility score \\
$\alpha,\beta,\gamma$ & Weights in MPC objective \\
$\mathbf{T}^{\text{lo}}_k,\mathbf{T}^{\text{osm}}_k$ & Local pose and OSM-aligned pose \\
\hline
\end{tabular}
\end{table}

The motorized LiDAR system produces raw points $\mathbf{r}^L_p$ in the LiDAR frame $\{L\}$, 
which are transformed into the world frame $\{W\}$ using the extrinsics 
$(\mathbf{R}^M_L,\mathbf{r}^M_L)$, the motor rotation $\mathbf{R}^B_M(\theta)$, 
and the base pose $(\mathbf{R}^W_B,\mathbf{r}^W_B)$. 
The motor state is parameterized by its angle $\theta$ and angular velocity $\omega$, 
and a sequence of speeds $\Omega=\{\omega_i\}$ is optimized in the control horizon. 
On the map side, $\mathcal{M}_{\text{OSM}}$ is constructed from extruded OSM footprints 
fused with DEM terrain, and each voxel is assigned a reliability weight $w_{\text{osm}}$. 
Given the current FoV $\mathcal{V}_k$ and history buffer $\mathcal{H}_k$, 
we build a refined local map and evaluate two metrics: 
$U(\theta)$, the A-optimal uncertainty of LiDAR odometry, 
and $P(\theta)$, the informativeness of OSM priors in that FoV. 
These scores enter the MPC cost with weights $(\alpha,\beta,\gamma)$. 
Finally, local odometry poses $\mathbf{T}^{\text{lo}}_k$ are anchored to 
global priors via OSM-aligned poses $\mathbf{T}^{\text{osm}}_k$ in a 
sliding-window graph. We summarized the proposed OSM-guided scanning control framework in Fig.\ref{fig:framework}, which will be detailed in the following sections.

\subsection{OSM+DEM Prior Map Construction}
To provide a lightweight global reference, we fuse OpenStreetMap (OSM) geometry with a Digital Elevation Model (DEM) from the Shuttle Radar Topography Mission (SRTM). OSM building footprints (closed ways) are extruded into 3D volumes using tagged heights or a default floor height of $3\,\mathrm{m}$/floor. The SRTM DEM is preprocessed and resampled in a uniform projected frame (e.g., UTM/ENU) to correct ground elevations and align the extruded footprints to the terrain. The resulting prior map is represented as a point cloud:
\begin{align}
\mathcal{M}_{\text{OSM}} = \{\,\mathbf{p}_i \in \mathbb{R}^3 \mid i=1,\dots,N \,\}.
\end{align}
Each point $\mathbf{p}_i$ originates either from the extruded OSM façades or from the SRTM-derived ground surface. This construction yields a prior in which façades are modeled more densely and the ground more sparsely, producing a map that is efficient to store and process while remaining geometrically compatible with LiDAR measurements.

\subsection{Panoramic Depth Image Generation for Control}

\begin{figure}[]
\includegraphics[width=\linewidth]{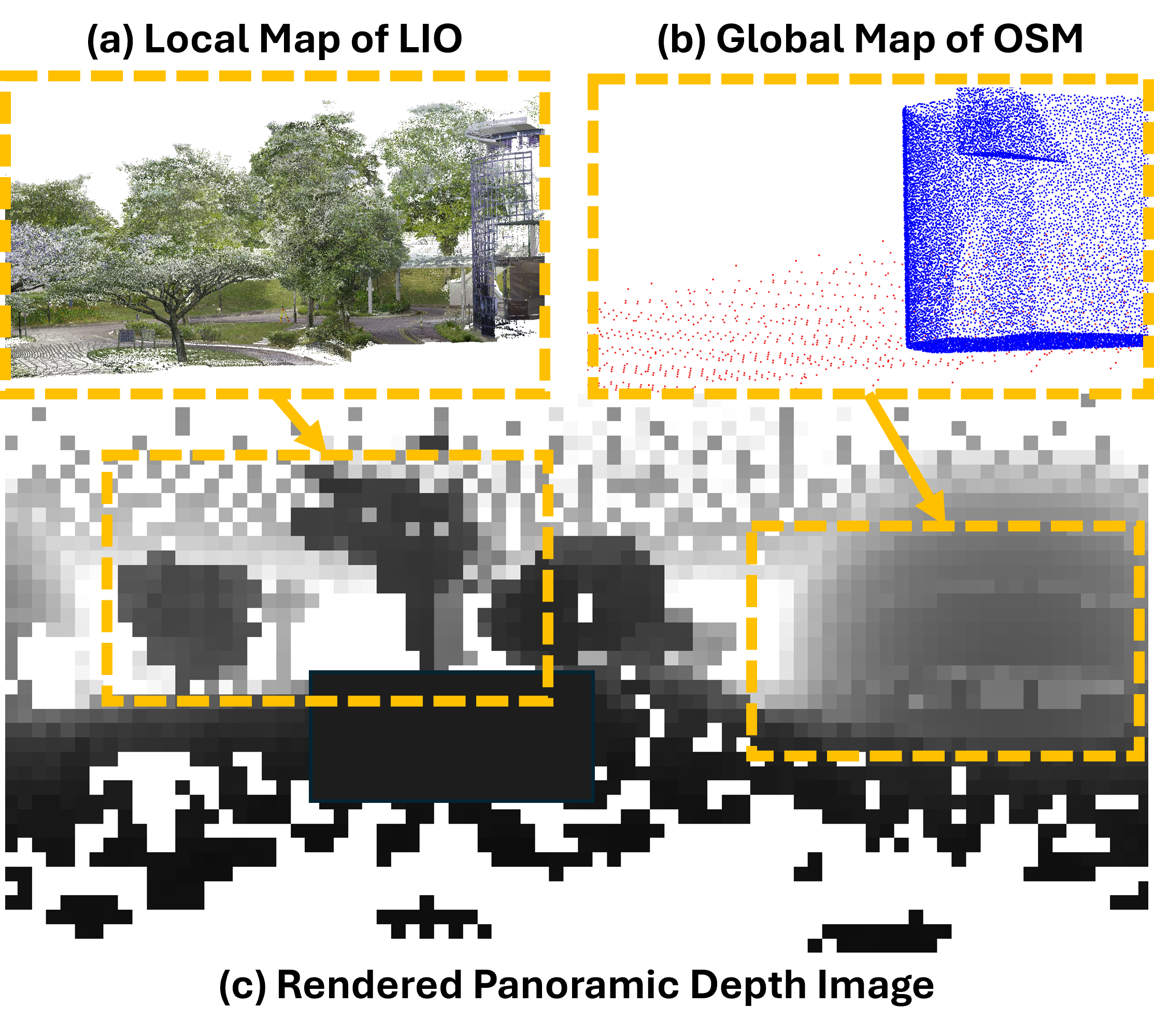}
    \captionof{figure}{Panoramic depth map for control. (a) $\mathcal{M}^{\text{loc}}_k$ for the local map maintained by LIO. (b) $\mathcal{M}^{\text{FoV}}_{\text{OSM}}$ be the subset of the OSM prior clipped to the current position. (c) The merged cloud $\mathcal{F}_k$ for control input.}
    \label{fig:pano_depth}
\end{figure}

We propose a panoramic projection pipeline that predicts future observability by fusing the LIO local map with the OSM prior, as shown in Fig.\ref{fig:pano_depth}.

At time $k$, let $\mathcal{M}^{\text{loc}}_k$ denote the local map maintained by LIO, and let $\mathcal{M}^{\text{FoV}}_{\text{OSM}}$ be the subset of the OSM prior clipped to the current position. These two sources are directly merged to form the input cloud:
\begin{align}
\mathcal{F}_k = \mathcal{M}^{\text{loc}}_k \cup \mathcal{M}^{\text{FoV}}_{\text{OSM}} .
\end{align}
After voxel downsampling, the refined cloud $\mathcal{F}_k$ is projected into a panoramic depth image $I_k^{\text{pan}}\in\mathbb{R}^{H\times W}$ using spherical projection:
\begin{align}
r &= \sqrt{x^2+y^2+z^2}, \\
\alpha &= \arctan2(y,x), \\
\beta &= \arcsin(z/r),\\
u &= \Big\lfloor \tfrac{\alpha-\alpha_{\min}}{\alpha_{\max}-\alpha_{\min}} (W-1)\Big\rfloor, \\
v &= \Big\lfloor \tfrac{\beta-\beta_{\min}}{\beta_{\max}-\beta_{\min}} (H-1)\Big\rfloor .
\end{align}
For each pixel $(u,v)$, only the minimum range $r$ is retained to handle occlusions.  
This panoramic representation encodes both already observed structures from LIO and global priors from OSM, serving as the basis for observability scoring and control.

\subsection{OSM-Guided Adaptive Motor Control}
We solve the allocation of angular speed directly with a receding-horizon MPC that balances odometry uncertainty, motion smoothness, and OSM prior utility. At each control step, the following optimization is solved:

\subsubsection{Objective}
\begin{align}
\min_{\{\omega_i\}} \quad 
J &= 
\alpha \sum_i U(\theta_i)
+ \beta \sum_i \big(\omega_i-\omega_{\mathrm{ref}}\big)^2
- \gamma \sum_i P(\theta_i).
\end{align}

\subsubsection{Constraints}
\begin{align}
\theta_{i+1} &= \theta_i + \omega_i \Delta t.
\end{align}

\subsubsection{Interpolation of scores}
The orientation-dependent scores are computed in a unified manner.  
For a candidate motor orientation $\theta$, point-to-plane residuals are linearized and accumulated into an information matrix \cite{li2025ua}:
\begin{align}
\Lambda(\theta) = \sum_i \mathbf{J}_i^\top \mathbf{J}_i, 
\qquad 
\mathbf{J}_i = \begin{bmatrix} [\mathbf{R}\mathbf{p}_i]_\times \mathbf{n}_i & \mathbf{n}_i \end{bmatrix},
\end{align}
where $\mathbf{p}_i$ and $\mathbf{n}_i$ are 3D points and normals from the map.  
The uncertainty score is then given by the A-optimal criterion
\begin{align}
S(\theta) = \mathrm{tr}\!\big(\Lambda(\theta)^{-1}\big).
\end{align}

When $\{\mathbf{p}_i,\mathbf{n}_i\}$ are drawn from the fused local map (LIO), the score is denoted $U(\theta)$;  
when they are drawn from the OSM prior, the score is denoted $P(\theta)$.  
Thus both metrics share the same formulation but reflect different data sources.

Since $U(\theta)$ and $P(\theta)$ vary smoothly with $\theta$, 
they are pre-sampled on a grid with step $\Delta\theta$ and interpolated during MPC:
\begin{align}
U(\theta_i) &\approx (1-\xi)\,U_k + \xi\,U_{k+1}, \\[4pt]
P(\theta_i) &\approx (1-\xi)\,P_k + \xi\,P_{k+1}, \\[4pt]
k &= \Big\lfloor \tfrac{\theta_i}{\Delta\theta} \Big\rfloor, \qquad
\xi = \tfrac{\theta_i}{\Delta\theta} - k .
\end{align}
This design ensures that both local observability and prior utility are evaluated consistently, 
with continuous values and efficient gradient computation for real-time control.

\subsubsection{Execution}
We warm-start the optimizer with the previous solution shifted by one step and run a fixed number of projected-gradient or SQP iterations within the time budget. The first control $\omega_0^\star$ is applied, the horizon shifts forward, and the procedure repeats at the LiDAR frame rate. This ensures smooth motor motion, guaranteed coverage, and more dwell time in FoV directions that are simultaneously informative for odometry and well supported by OSM priors.

\subsection{OSM-Aided Global Fusion}
We anchor local odometry to the OSM prior at a low rate to constrain long-term drift while preserving the consistency of IMU-aided estimation. Let $\mathbf{T}^{\text{lo}}_k\!\in\!\mathrm{SE}(3)$ be the local pose from FAST-LIO \cite{Xu2020FASTLIOAF} at time $k$, and let $\mathbf{T}^{\text{osm}}_k$ be a scan-to-map alignment against the OSM+DEM prior $\mathcal{M}_{\text{OSM}}$ (point-to-line/plane ICP with robust gating and reliability $w_{\text{osm}}$). The OSM residual is
\begin{align}
\boldsymbol{\xi}_k &= \mathrm{Log}\!\left((\mathbf{T}^{\text{osm}}_k)^{-1}\mathbf{T}^{\text{lo}}_k\right) \in \mathfrak{se}(3), \\
\boldsymbol{\xi}_k &= \begin{bmatrix}\mathbf{v}_k \\ \mathbf{w}_k\end{bmatrix}, \qquad
\mathbf{v}_k,\mathbf{w}_k \in \mathbb{R}^3 .
\end{align}

\subsubsection{Sliding-window factor and gating}
We maintain a fixed-size window $\mathcal{W}$ of the most recent successful OSM alignments. The fused objective over $\mathcal{W}$ is
\begin{align}
\min_{\{\mathbf{T}^{\text{lo}}_k\}} \;\; \sum_{k\in\mathcal{W}} \mathbf{r}_k^\top \boldsymbol{\Sigma}^{-1}\mathbf{r}_k ,
\end{align}
where $\mathbf{r}_k$ stacks LO/IMU factors (internal to FAST-LIO) and the OSM factor built from $\boldsymbol{\xi}_k$. To reject poor alignments we use a normalized error and a hard gate:
\begin{align}
e_{t,k} &= \|\mathbf{v}_k\|/\sigma_t, \qquad
e_{r,k} = \|\mathbf{w}_k\|/\sigma_r, \\
e_k &= \sqrt{\tfrac{e_{t,k}^2 + \lambda\, e_{r,k}^2}{1+\lambda}}, \qquad
\text{accept if } e_k \le \tau_g .
\end{align}
Here $\sigma_t,\sigma_r$ scale translation/rotation, and $\lambda$ balances them. 

\subsubsection{Huber weighting and robust aggregation}
For $k\!\in\!\mathcal{W}$, a scalar weight is assigned by the Huber kernel (unit threshold):
\begin{align}
w_H(e_k) &= 
\begin{cases}
1, & |e_k|\le 1,\\
1/|e_k|, & |e_k|>1 .
\end{cases}
\end{align}
The effective weight is $w_k = w_{\text{osm},k}\, w_H(e_k)$. We aggregate directions and magnitudes of translation and rotation separately:
\begin{align}
\mathbf{d}_t &= \frac{\sum_{k\in\mathcal{W}} w_k\, \mathbf{v}_k}{\big\|\sum_{k\in\mathcal{W}} w_k\, \mathbf{v}_k\big\|}, \qquad
\mathbf{d}_r = \frac{\sum_{k\in\mathcal{W}} w_k\, \mathbf{w}_k}{\big\|\sum_{k\in\mathcal{W}} w_k\, \mathbf{w}_k\big\|}, \\
m_t &= \frac{\sum_{k\in\mathcal{W}} w_k \,\|\mathbf{v}_k\|}{\sum_{k\in\mathcal{W}} w_k}, \qquad
m_r = \frac{\sum_{k\in\mathcal{W}} w_k \,\|\mathbf{w}_k\|}{\sum_{k\in\mathcal{W}} w_k}.
\end{align}
Saturation prevents aggressive jumps:
\begin{align}
\mathbf{s}_t &= \min(m_t,\rho_t)\,\mathbf{d}_t, \qquad
\mathbf{s}_r = \min(m_r,\rho_r)\,\mathbf{d}_r .
\end{align}

\subsubsection{Sparse, bounded feedback}
A small-step correction in $\mathfrak{se}(3)$ is injected at a low rate (e.g., every $L$ frames):
\begin{align}
\boldsymbol{\xi}_{\text{upd}} &= \eta\,
\begin{bmatrix}
\mathbf{s}_t \\
\mathbf{s}_r
\end{bmatrix}, \qquad
\mathbf{T}_{\text{corr}} \leftarrow \exp(\boldsymbol{\xi}_{\text{upd}})\, \mathbf{T}_{\text{corr}} .
\end{align}
The correction is applied as a bounded bias on the LO trajectory initialization rather than overwriting optimized states, preserving IMU preintegration consistency. Graph fusion and feedback are executed only when ICP converges; the target map remains in the OSM world frame.

In this subsection, $\mathbf{T}^{\text{lo}}_k$ denotes the local odometry pose estimated by FAST-LIO\cite{Xu2020FASTLIOAF}, while $\mathbf{T}^{\text{osm}}_k$ is the pose aligned to the OSM prior. Their relative difference is expressed as the residual $\boldsymbol{\xi}_k \in \mathfrak{se}(3)$, which is decomposed into a translational component $\mathbf{v}_k$ and a rotational component $\mathbf{w}_k$. The residuals are normalized using $\sigma_t$ and $\sigma_r$ for translation and rotation, respectively, and balanced by a factor $\lambda$. A gating threshold $\tau_g$ is used to reject outliers, while the reliability of each OSM factor is encoded in $w{\text{osm},k}$. Robust weighting is applied through the Huber kernel $w_H$. To avoid aggressive corrections, the aggregated translation and rotation steps are saturated by limits $\rho_t$ and $\rho_r$. The final correction update is scaled by a step size $\eta$ and injected only every $L$ frames to maintain system stability.

\section{Experiments}

\subsection{Simulation Environment}
We extend MARSIM \cite{kong2023} to support motor commands, scanning patterns, and controllable OSM completeness. A ROS-based pipeline replays ground-truth trajectories, simulates LiDAR rays under commanded rotations, and feeds the LO and motor control modules.  
The evaluation metric is the Absolute Trajectory Error (ATE). This setup provides a principled testbed for evaluating adaptive scanning policies under varying prior availability.

\subsection{Evaluation on the Campus Dataset}
\begin{figure}[t]
\includegraphics[width=\linewidth]{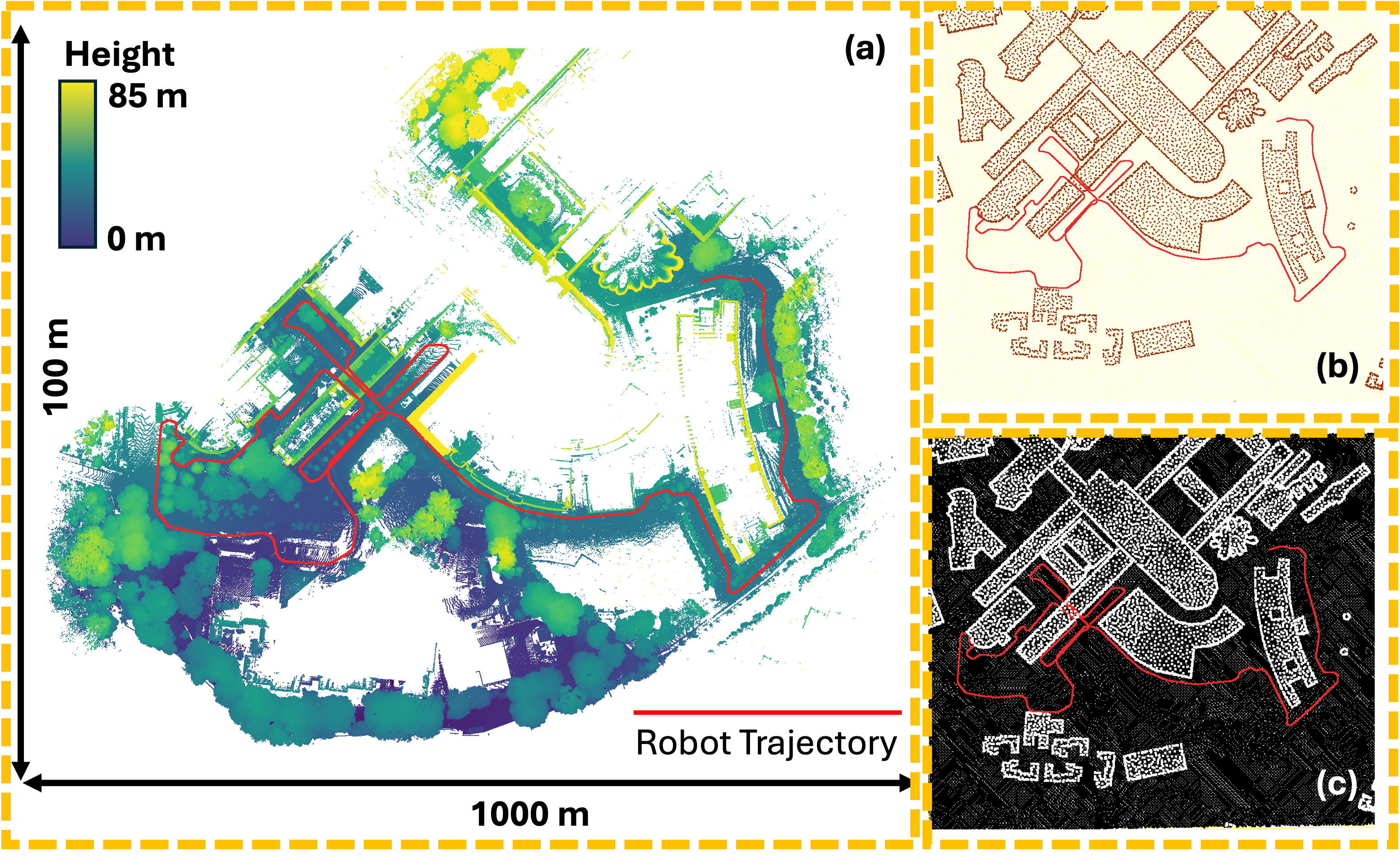}
\caption{Campus simulation setup. 
(a) Ground-truth 3D scene with height coloring and trajectory. 
(b) OSM footprints for the area. 
(c) OSM+DEM prior constructed by extruding OSM and aligning with SRTM elevation.}
\label{fig:simu_exp}
\end{figure}

\begin{figure}[t]
\includegraphics[width=\linewidth]{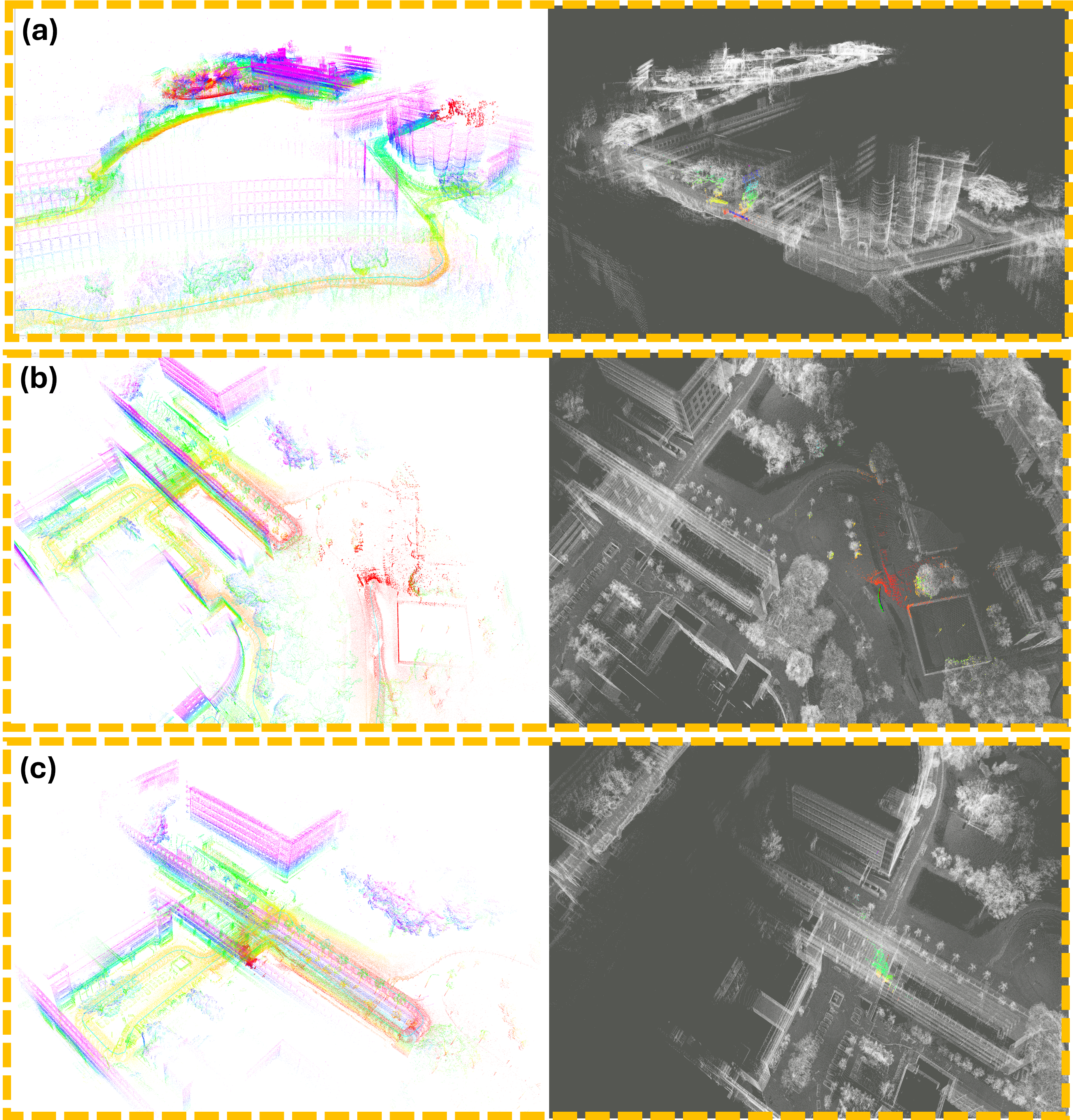}
\caption{Simulation in MARSIM. 
Snapshots of the active motorized system: left—localization view; right—rendering in the high-accuracy ground-truth cloud.}
\label{fig:simu_marsim}
\end{figure}

We evaluate the proposed method in a campus-scale simulation built from a high-fidelity 3D model, with OSM footprints extruded and aligned to SRTM elevation (Fig.~\ref{fig:simu_exp}). 
The environment features long corridors, open squares, and dense building clusters. 
We then run different motor control strategies in the MARSIM simulator (Fig.~\ref{fig:simu_marsim}) to assess their impact on localization\cite{kong2023}.

\paragraph*{Compared strategies}
\begin{itemize}
  \item \textbf{0\,rad/s (static):} LiDAR is fixed, providing limited FoV coverage.
  \item \textbf{3\,rad/s (constant low speed):} Uniform panoramic scanning at a low rate.
  \item \textbf{9\,rad/s (constant fast speed):} Uniform panoramic scanning at a high rate.
  \item \textbf{Proposed (OSM-guided):} Adaptive scanning that allocates dwell time to FoV sectors with high odometry observability and informative OSM structures.
\end{itemize}

\begin{figure}[]
\centering
\includegraphics[width=\linewidth]{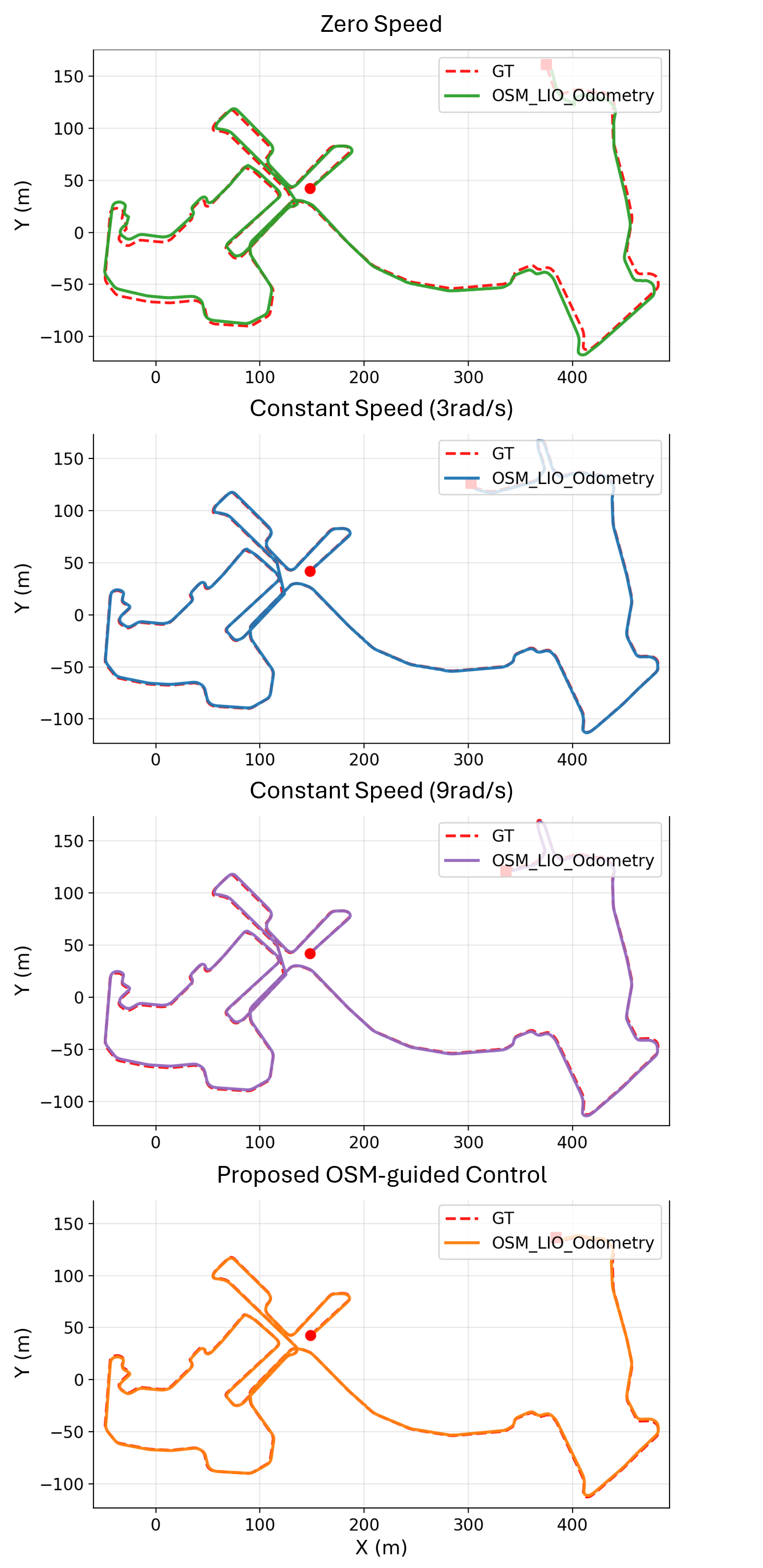}
\caption{Estimated trajectories under different motor control strategies.}
\label{fig:trajectories}
\end{figure}

\begin{figure}[]
\centering
\includegraphics[width=\linewidth]{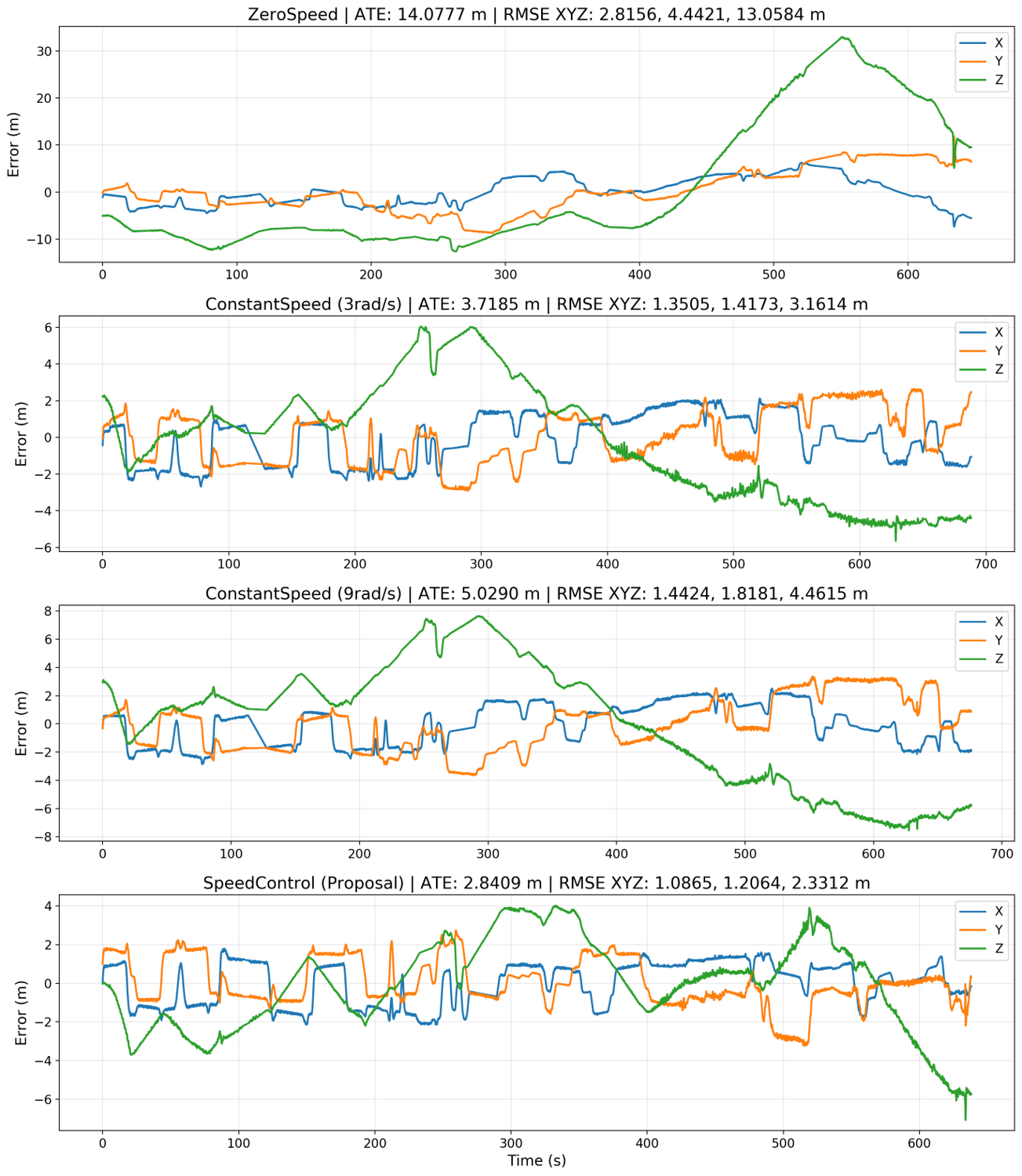}
\caption{Position error over time (X, Y, Z) for each strategy. 
Note the dominant Z-axis drift in static and constant-speed cases, which is mitigated by the proposed controller.}
\label{fig:errors}
\end{figure}

Trajectory accuracy is reported for all settings.
Fig.~\ref{fig:trajectories} compares the estimated trajectories, while Fig.~\ref{fig:errors} illustrates the per-axis error evolution over time.

In the \textbf{static case (0,rad/s)}, the LiDAR remains fixed and the effective FoV is extremely limited. As a result, large portions of the environment are never observed, leading to accumulated drift along the 2,km trajectory. The error curves reveal severe growth in the vertical (Z) component, exceeding 10,m, which highlights the system’s inability to constrain altitude without sufficient structural features.

With \textbf{constant-speed rotation}, coverage improves because the sensor sweeps panoramically. At a low rate of 3,rad/s, the denser sampling per orientation provides some benefit, but drift still emerges in open or feature-sparse regions, especially along the Z axis. At a higher rate of 9,rad/s, the LiDAR distributes effort uniformly but the effective point density per sweep decreases, making the odometry more sensitive to noise and degeneracy. This trade-off explains why the 3,rad/s setting achieves lower APE than 9,rad/s, although both constant-speed strategies still exhibit wasted scanning effort in directions that contribute little to localization.

In contrast, the \textbf{proposed OSM-guided controller} explicitly balances local observability and global priors. The motor slows down in FoV sectors that both enhance odometry constraints and coincide with OSM building façades, while maintaining sufficient angular coverage. This adaptive redistribution of scan time reduces wasted effort in uninformative regions and reinforces alignment in map-supported areas. Quantitatively, the proposed method achieves the lowest APE (2.84,m) and RMSE across all axes, with a particularly notable reduction of vertical error compared to baselines. Qualitatively, the trajectory in Fig.~\ref{fig:trajectories} stays close to the ground-truth path without the large drifts observed in static or constant-speed cases.

These results demonstrate that coupling OSM priors with adaptive scanning yields robustness and efficiency beyond constant-speed baselines. The controller not only minimizes overall trajectory drift but also improves stability in the most challenging dimensions, validating the effectiveness of the proposed design.

\begin{table}[t]
\centering
\caption{Absolute Pose Error (APE) on the campus trajectory.}
\label{tab:ape_results}
\begin{tabular}{lcc}
\hline
Method & Mean APE [m] & RMSE APE in (X, Y, Z) [m] \\
\hline
Static (0\,rad/s)         & 14.078 & (2.816, 4.442, 13.058) \\
Constant (3\,rad/s)       & 3.719  & (1.351, 1.417, 3.161) \\
Constant (9\,rad/s)       & 5.029  & (1.442, 1.818, 4.462) \\
Proposed (OSM-guided)     & \textbf{2.841} & \textbf{(1.086, 1.206, 2.331)} \\
\hline
\end{tabular}
\end{table}

These results confirm the benefit of coupling map priors with adaptive scanning: the proposed controller achieves the lowest APE and per-axis RMSE, particularly reducing the vertical (Z) error where constant-speed baselines are most affected, while preserving overall scan coverage.

\subsection{Discussion on OSM Data Missing}

\begin{figure}[]
\centering
\includegraphics[width=0.9\linewidth]{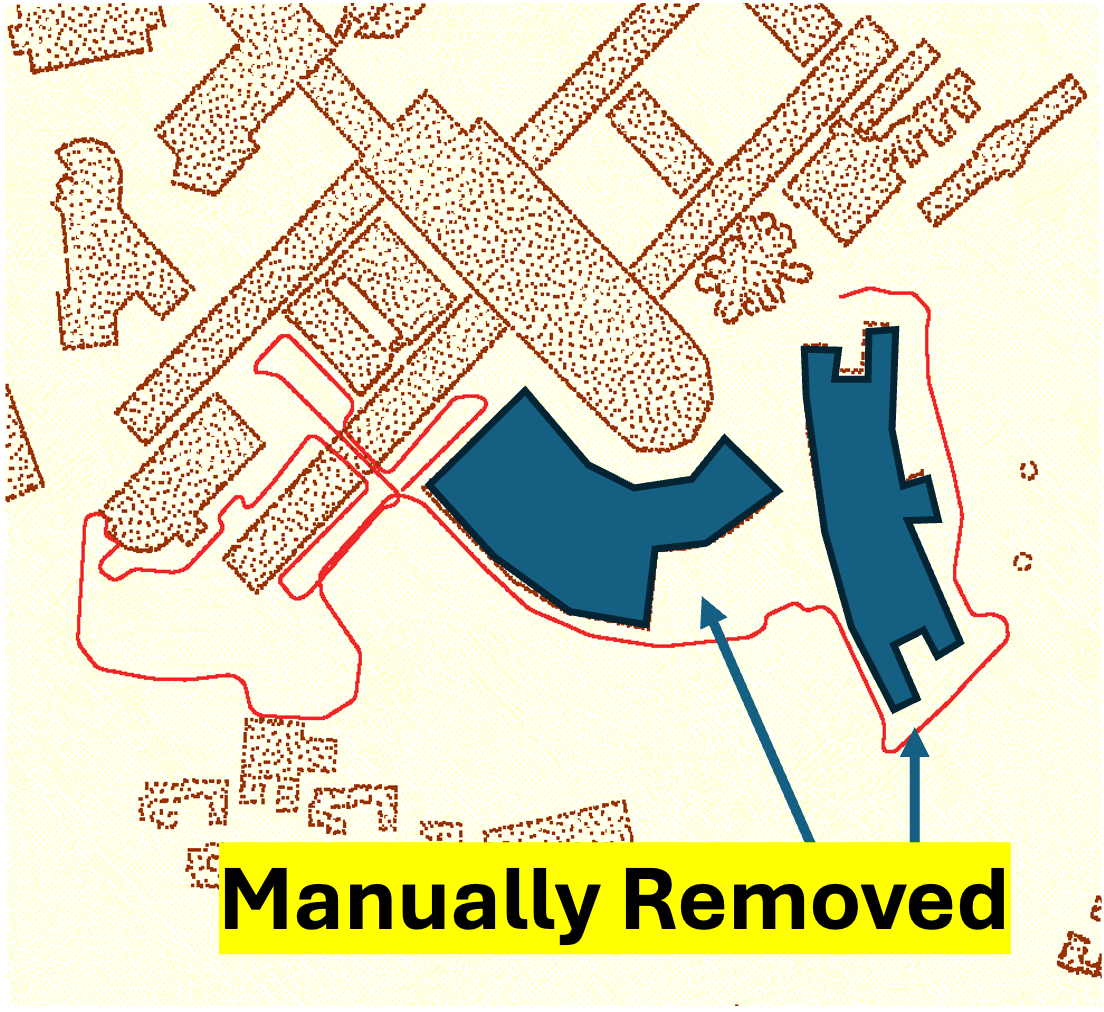}
\caption{Manually removed data in the raw OSM to simulate data missing.}
\label{fig:manually_removed}
\end{figure}

\begin{figure}[]
\centering
\includegraphics[width=\linewidth]{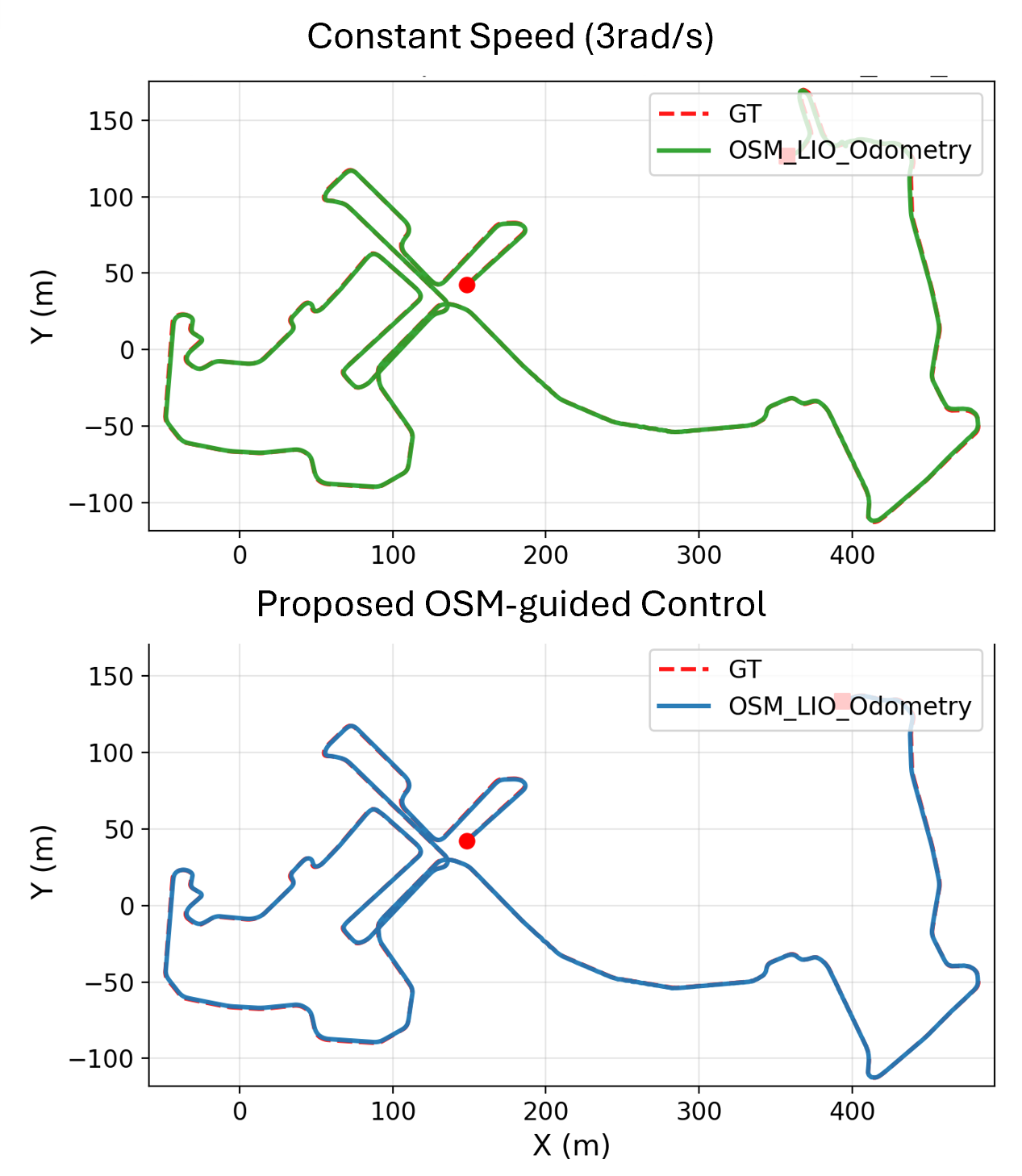}
\caption{Estimated trajectories under different motor control strategies on the map with missing data.}
\label{fig:missing_trajectories}
\end{figure}

\begin{figure}[]
\centering
\includegraphics[width=\linewidth]{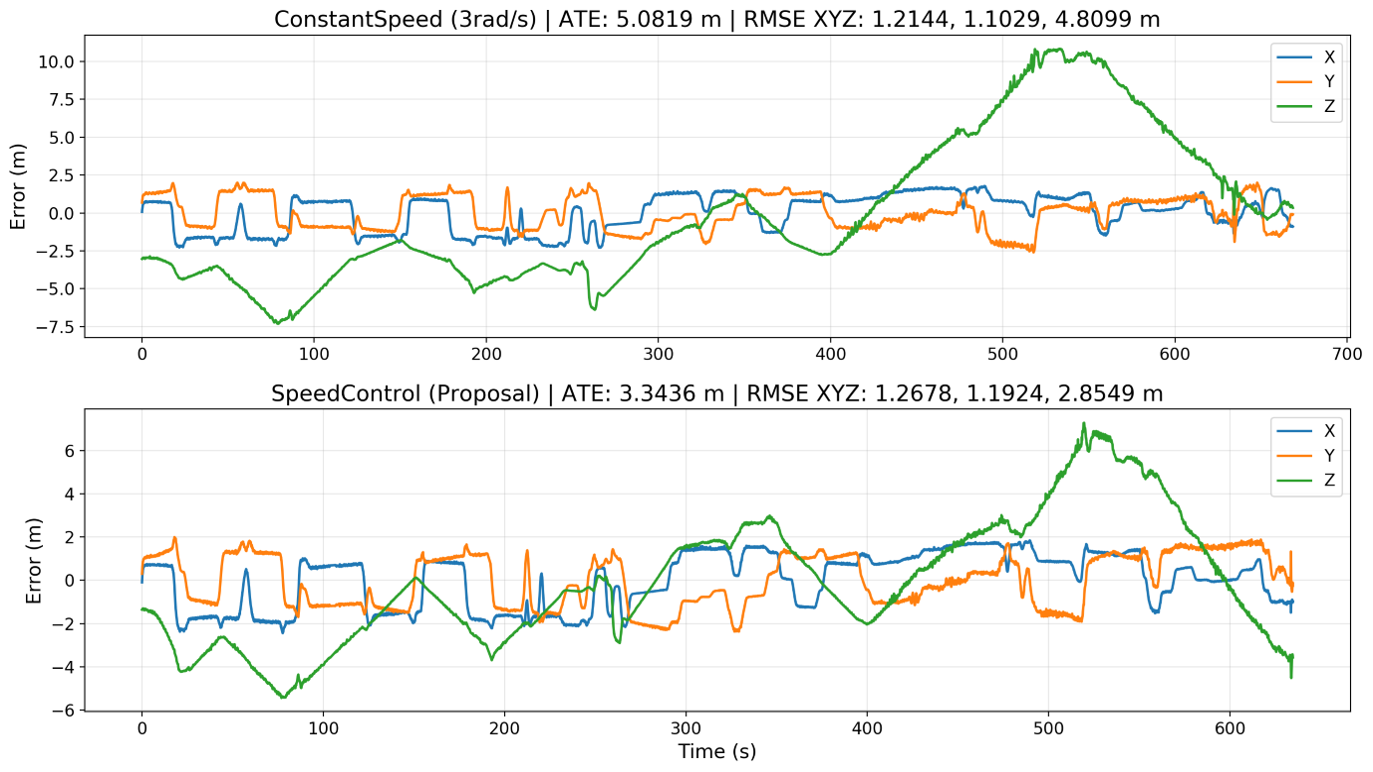}
\caption{Position error over time (X, Y, Z) for each strategy on the map with missing data.}
\label{fig:missing_errors}
\end{figure}

To analyze robustness under incomplete priors, we simulate OSM data deficiency by selectively removing building footprints from the base map as shown in Fig. \ref{fig:manually_removed}. We then compare the proposed OSM-guided controller with the constant-speed baseline at 3\,rad/s, as shown in Fig.~\ref{fig:missing_trajectories} and Fig.~\ref{fig:missing_errors}.  

The results reveal that the baseline method is highly sensitive to missing priors. Without reliable map support, the error along the Z axis grows significantly, and the trajectory accuracy degrades, with a mean APE of 5.08\,m. In contrast, our OSM-guided strategy maintains stable localization, reducing the mean APE to 3.34\,m. The per-axis RMSE also demonstrates consistent improvement, especially in the vertical direction, where the proposed method limits drift by reallocating scanning effort toward regions with remaining OSM structures.  

These findings highlight an important property of our framework: even when OSM data are incomplete, the controller adaptively exploits both local observability and the available priors to stabilize odometry. Quantitatively, the proposed approach reduces APE by 42\% compared to the 3\,rad/s baseline, confirming that coupling OSM priors with adaptive motor control provides robustness against map sparsity and incompleteness.

\section{Conclusion}
In this work, we presented an adaptive LiDAR scanning framework that leverages OpenStreetMap (OSM) priors to improve localization robustness in large-scale environments. Unlike constant-speed rotation, which allocates scanning effort uniformly, our method integrates local odometry observability with the spatial distribution of OSM features to guide motor control. This design enables the sensor to slow down in informative directions while maintaining overall coverage.  

We demonstrated that the proposed OSM-guided strategy can be implemented in real time using a model predictive control formulation with efficient interpolation of observability scores. Experiments on a campus-scale trajectory showed that our method consistently reduces absolute pose error compared to static and constant-speed baselines, while preserving scan completeness.  

Future work will extend the framework to more complex map sources and multi-modal settings, such as coupling LiDAR with camera imagery or semantic maps. Another direction is to explore long-term deployment with online map updates, where the scanning policy adapts not only to the current FoV but also to dynamically changing map priors.

\bibliographystyle{ieeetr}
\bibliography{refs} 
\end{document}